\documentclass[sigconf,nonacm]{acmart}
\usepackage{tikz}
\usetikzlibrary{arrows.meta,backgrounds,calc,fit,positioning}
\definecolor{dceoblue}{HTML}{3B73B9}
\definecolor{dceopurple}{HTML}{7157A8}
\definecolor{dceogreen}{HTML}{3B8C73}
\definecolor{dceoorange}{HTML}{D17A35}
\definecolor{dceocyan}{HTML}{2C7F9E}
\definecolor{dceorose}{HTML}{B55473}
\AtBeginDocument{%
  }

\setcopyright{none}
\begin{document}

\title{DCEO: Direct Causal Effect Optimization for Long-Term User Value Modeling in E-commerce Search}

\author{Junzhao Zhang}
\email{junzhao.zjz@alibaba-inc.com}
\affiliation{%
  \institution{Taobao \& Tmall Group of Alibaba}
  \city{Hangzhou}
  \state{Zhejiang}
  \country{China}
}

\author{Tao Zhang}
\affiliation{%
  \institution{Taobao \& Tmall Group of Alibaba}
  \city{Beijing}
  \country{China}
}
\email{quen.zt@alibaba-inc.com}

\author{Liren Yu}
\affiliation{%
  \institution{Taobao \& Tmall Group of Alibaba}
  \city{Hangzhou}
  \state{Zhejiang}
  \country{China}
}
\email{yuliren.ylr@taobao.com}

\author{Feiyi Dong}
\affiliation{%
  \institution{Taobao \& Tmall Group of Alibaba}
  \city{Hangzhou}
  \state{Zhejiang}
  \country{China}
}
\email{dongfeiyi.dfy@taobao.com}

\author{Zhixuan Zhang}
\affiliation{%
  \institution{Taobao \& Tmall Group of Alibaba}
  \city{Hangzhou}
  \state{Zhejiang}
  \country{China}
}
\email{zhibing.zzx@taobao.com}

\author{Dan Ou}
\affiliation{%
  \institution{Taobao \& Tmall Group of Alibaba}
  \city{Hangzhou}
  \state{Zhejiang}
  \country{China}
}
\email{oudan.od@taobao.com}

\author{Haihong Tang}
\affiliation{%
  \institution{Taobao \& Tmall Group of Alibaba}
  \city{Hangzhou}
  \state{Zhejiang}
  \country{China}
}
\email{piaoxue@taobao.com}



\begin{abstract}
Industrial e-commerce search systems ultimately aim to optimize the user-level long-term objective, such as $n$-day cumulative purchases or gross merchandise value (GMV) per user. However, such objectives are defined at the user level, whereas search ranking is based on item-level scores within each request. Existing methods typically bridge this granularity gap through manually designed multi-objective fusion, where predicted scores for multiple item-level objectives, such as click, cart, purchase, and transaction value, are combined into a ranking score that serves as a proxy for the ultimate objective. Such hand-crafted fusion schemes rely on a small set of manually tuned weights, limiting fine-grained personalization and leading to suboptimal alignment with the ultimate objective. In this paper, we propose DCEO (Direct Causal Effect Optimization), a data-driven framework for learning item-level proxy scores that are better aligned with the ultimate objective. We first aggregate the item-level proxy scores into a user-level proxy metric and quantify its alignment with the ultimate objective using a relative causal effect. We then develop an actor-critic framework, where the critic estimates the ultimate objective for a given user-level proxy metric, and the actor dynamically generates context-dependent fusion weights over multiple objectives to construct the item-level proxy scores and is trained to directly optimize the relative causal effect. Extensive offline experiments and analyses demonstrate the effectiveness and interpretability of DCEO. In addition, DCEO has been deployed in a large-scale industrial e-commerce search system, outperforming the conventional GMV proxy by 0.36\% in GMV in a 41-day online A/B test.
\end{abstract}




\maketitle

\section{Introduction}
\label{sec:introduction}

E-commerce search systems ultimately aim to optimize user-level long-term objectives, such as cumulative purchases or gross merchandise value (GMV) over an $n$-day window. For each request, the ranking stage assigns a score to each candidate item and orders the items by their scores \cite{hu2018rlrank,liu2017cascade}. However, these ranking scores are defined at the item level, whereas the ultimate objective summarizes a user's outcome over multiple requests and impressions. This granularity gap prevents the user-level objective from directly serving as an item-level training label, making it challenging to optimize long-term user value through item ranking.

Industrial ranking systems commonly mitigate this granularity gap with manually designed multi-objective fusion. Upstream models predict multiple item-level objectives, such as click, cart, purchase, and transaction value \cite{santu2017ecommerce}, and the fusion module combines their predicted scores into a ranking score \cite{zhao2019multitaskranking}. Practitioners improve the alignment between this ranking score and the user-level long-term objective by adding new signals or tuning fusion parameters through repeated online A/B tests. This process has two limitations: the small set of globally shared parameters provides limited personalization across users and requests, and repeated online experiments are costly and time-consuming. We therefore seek to develop a data-driven method for learning a context-dependent item-level proxy score directly from the user-level ultimate objective and incorporating it into the existing multi-objective fusion formula.

The key to learning such a proxy score is to define its alignment with the user-level ultimate objective. We evaluate the item-level proxy scores through their aggregated user-level proxy metric, which has the same granularity as the ultimate objective. A natural approach is to optimize the predictive association between the proxy metric and the ultimate objective. However, predictive association does not imply that increasing the proxy metric through ranking will produce a large improvement in the ultimate objective \cite{vanderweele2013surrogate}. Since adding the proxy score to the fusion formula is intended to increase the proxy metric, we instead evaluate how much the ultimate objective improves under a given relative increase in the proxy metric. This motivates optimizing the relative causal effect of the proxy metric on the ultimate objective.

In this paper, we propose \emph{Direct Causal Effect Optimization} (DCEO), a data-driven actor-critic framework that directly optimizes the relative causal effect of the proxy metric on the ultimate objective. The actor generates context-dependent weights over selected upstream predicted scores and combines them into an item-level proxy score. DCEO aggregates the proxy scores over a user's impressions and calibrates the aggregate to a reference impression count, yielding a user-level proxy metric. The critic predicts the ultimate objective from the user features and proxy metric, and its predictions before and after a relative increase in the proxy metric provide the training signal for the actor. By maximizing the estimated causal effect, the actor learns a personalized item-level proxy score directly from user-level supervision.

DCEO separates offline training from online serving. The critic and calibrated user-level aggregation are used only during training, while only the actor is deployed online. Given user and request features available online and selected upstream predicted scores, the actor generates an item-level proxy score. This score is added as a new term to the existing multi-objective fusion formula, while the original fusion weights and score components remain unchanged. In this way, DCEO learns an item-level score from user-level long-term supervision for direct use in online ranking.

The main contributions of this work are summarized as follows:
\begin{itemize}
\item We formulate learning an item-level proxy score from a user-level long-term objective as a cross-granularity optimization problem. We characterize the alignment between the proxy score and the ultimate objective using the relative causal effect of an aggregated user-level proxy metric on the ultimate objective.
\item We propose DCEO, a data-driven actor-critic framework that learns context-dependent proxy scores by directly optimizing the critic-estimated relative causal effect. Calibrated user-level aggregation connects item-level actor outputs with user-level supervision, enabling end-to-end learning of the proxy score.
\item Extensive offline experiments and analyses demonstrate the effectiveness and interpretability of DCEO. In a 41-day online A/B test in a large-scale industrial e-commerce search system, DCEO outperforms the conventional GMV proxy by 0.36\% in GMV.
\end{itemize}

\section{Related Work}
\subsection{Long-Term User Value Modeling}

Recent studies have increasingly shifted from optimizing item-level short-term feedback to optimizing user-level long-term objectives. However, these objectives are typically delayed, sparse, and defined at a coarser granularity than item-level ranking, making it difficult to derive effective item-level training signals. To address this challenge, one line of research constructs proxy objectives. Behavior-based methods identify signals associated with future visits and retention from interaction logs \cite{wang2022surrogate,wang2026downstream}, while human-feedback methods use return-intent surveys or preference comparisons to capture aspects of long-term user experience that may not be reflected by immediate behavior \cite{bakhshi2026retentive,xue2023prefrec}. AURO converts session-level return time into a terminal retention reward and propagates it to preceding ranking decisions through sequential policy learning \cite{xue2025auro}. Future Impact Decomposition more explicitly bridges the granularity gap by allocating request-level future value to items according to immediate feedback or learned weights, thereby constructing item-level learning targets \cite{wang2024future}. These methods derive tractable item-level supervision by constructing proxy objectives or decomposing future value, rather than directly learning an item-level proxy score from the ultimate objective.

Another line of research directly incorporates the ultimate objective into supervision without explicitly decomposing it into item-level targets. IURO learns item-level retention scores by using attention-based aggregation to connect item-level representations with a user-level retention outcome and introduces manually designed auxiliary attribution tasks for interpretability \cite{ding2023interpretable}. However, the attention weights that establish this connection during training are unavailable in the same form during online serving, resulting in training-serving inconsistency. Despite their different formulations, these methods primarily learn predictive associations: a signal that accurately predicts the ultimate objective does not necessarily imply that increasing it through ranking will improve that objective. In contrast, DCEO aggregates item-level proxy scores into a user-level proxy metric and, to the best of our knowledge, is the first framework to learn the item-level proxy score from the user-level ultimate objective by directly optimizing the estimated relative causal effect of the proxy metric on the ultimate objective.

\subsection{Multi-Objective Fusion}

Multi-objective fusion combines heterogeneous predicted scores into a single ranking score. Industrial systems traditionally rely on hand-crafted formulas with globally shared parameters, which require repeated online tuning and provide limited personalization. One line of work learns context-dependent fusion parameters through reward or policy optimization. Value-aware recommendation maps user actions to monetized rewards and optimizes their aggregate economic value \cite{pei2019valueaware}. BatchRL-MTF formulates fusion as a session-level Markov decision process and uses offline reinforcement learning to optimize a reward constructed from user stickiness and activeness \cite{zhang2022batchrlmtf}. More recent methods improve policy learning under sparse industrial feedback: GRADE combines group-relative optimization with structured exploration over fusion weights \cite{hong2026grade}, while SaFRO constructs a query-level satisfaction reward and employs dual-relative policy optimization for short-video search \cite{zhou2026safro}. These methods improve personalized fusion, but their optimization targets are monetized behaviors or manually constructed satisfaction and retention rewards rather than the user-level ultimate objective itself.

Another line of work replaces the hand-crafted fusion formula with a trainable ensemble model. Pantheon inherits hidden representations from upstream task models and uses iterative Pareto optimization to balance multiple objectives \cite{cao2025pantheon}. EMER organizes candidates at the request level and uses a Transformer-based listwise model with self-evolving multi-objective supervision \cite{he2025emer}. UMRE learns personalized monotonic transformations of upstream predicted scores before combining them with a lightweight ensemble model \cite{xu2025umre}. These approaches substantially increase the capacity and personalization of the fusion module, but they are trained primarily with item-level behavior labels and task-specific metrics. They therefore do not directly address the granularity gap between item-level ranking and user-level long-term objectives.

Whole-page optimization introduces long-term causal evidence by estimating the effects of page-level quality metrics on delayed user feedbacks from quasi-experimental data \cite{lahiri2026wholepage}. However, the estimated effects are used to construct fixed weights for a whole-page objective rather than serving as the optimization target for learning an item-level proxy. In contrast, DCEO learns a personalized proxy score from the user-level ultimate objective. It connects item-level actor outputs with user-level supervision through calibrated aggregation and directly optimizes the relative causal effect of the resulting proxy metric on the ultimate objective.

\begin{figure*}[t]
  \centering
  \resizebox{\textwidth}{!}{%
  \begin{tikzpicture}[
    font=\sffamily\small,
    >=Latex,
    node distance=5mm and 8mm,
    input/.style={draw=gray!65, fill=gray!8, rounded corners=1.5pt,
      minimum width=24mm, minimum height=10mm, align=center, inner sep=3pt},
    actor/.style={draw=dceopurple!85!black, fill=dceopurple!13,
      rounded corners=2pt, minimum width=28mm, minimum height=14mm,
      align=center, very thick, inner sep=4pt},
    proxy_score/.style={draw=dceoblue!85!black, fill=dceoblue!12,
      rounded corners=2pt, minimum width=30mm, minimum height=14mm,
      align=center, thick, inner sep=4pt},
    proxy_metric/.style={draw=dceogreen!75!black, fill=dceogreen!13,
      rounded corners=2pt, minimum width=32mm, minimum height=14mm,
      align=center, thick, inner sep=4pt},
    critic/.style={draw=dceoorange!85!black, fill=dceoorange!14,
      rounded corners=2pt, minimum width=30mm, minimum height=14mm,
      align=center, very thick, inner sep=4pt},
    loss/.style={draw=gray!70, fill=white, rounded corners=2pt,
      minimum width=27mm, minimum height=11mm, align=center, thick,
      inner sep=3pt},
    serving/.style={draw=dceoblue!75!black, fill=white, rounded corners=2pt,
      minimum width=30mm, minimum height=14mm, align=center, thick,
      inner sep=3pt},
    rank_score/.style={draw=dceocyan!80!black, fill=dceocyan!13,
      rounded corners=2pt, minimum width=30mm, minimum height=14mm,
      align=center, thick, inner sep=3pt},
    ranking_output/.style={draw=dceorose!85!black, fill=dceorose!14,
      rounded corners=2pt, minimum width=30mm, minimum height=14mm,
      align=center, thick, inner sep=3pt},
    flow/.style={-{Latex[length=2.2mm]}, thick, draw=gray!80},
    grad/.style={-{Latex[length=2.2mm]}, thick, dashed,
      draw=dceopurple!85!black},
    aux/.style={-{Latex[length=2mm]}, semithick, dashed, draw=gray!65},
    group/.style={draw=gray!55, rounded corners=4pt, line width=0.7pt},
    tag/.style={font=\sffamily\bfseries\small, text=gray!25!black,
      fill=white, inner sep=2pt}
  ]

  \node[input] (context) {User/request features\\$\mathbf z_u,\mathbf z_{ur}$};
  \node[input, below=12mm of context] (preds) {Selected upstream\\predicted scores $\mathbf v_{uri}$};
  \node[actor, right=10mm of context] (actor) {Actor\\$\mathbf w_{ur}=f_\theta(\mathbf z_u,\mathbf z_{ur})$};
  \node[proxy_score, right=10mm of actor] (proxy_score) {Proxy score\\$p_{uri}=\sum_{m=1}^M w_{ur}^m v_{uri}^m$};
  \node[proxy_metric, right=10mm of proxy_score] (proxy_metric) {Proxy metric\\$P_u=\operatorname{Aggregation}_{(r,i)\in\mathcal I_u}(p_{uri})$};
  \node[critic, right=10mm of proxy_metric] (critic) {Critic\\$h_\phi(\mathbf z_u,P_u)$};
  \node[loss, above=5mm of actor] (actorloss) {Actor loss\\$\mathcal L_{\mathrm{actor}}=\mathcal{L}_{\mathrm{CE}}+\alpha\mathcal{L}_{\mathrm{CNR}}$};
  \node[loss, above=5mm of critic] (criticloss) {Critic loss\\$\mathcal L_{\mathrm{critic}}$};

  \draw[flow] (context.east) -- (actor.west);
  \draw[flow] (actor.east) -- (proxy_score.west);
  \draw[flow] (preds.east) -| (proxy_score.south);
  \draw[flow] (proxy_score.east) -- (proxy_metric.west);
  \draw[flow] (proxy_metric.east) -- (critic.west);
  \draw[flow] (context.south) -- ++(0,-10mm) -| (critic.south);
  \draw[grad] (actorloss.south) -- (actor.north);
  \draw[grad] (criticloss.south) -- (critic.north);

  \begin{scope}[on background layer]
    \node[group, fit=(context)(preds)(actor)(proxy_score)(proxy_metric)(critic)
      (actorloss)(criticloss),
      inner sep=7mm] (trainingbox) {};
  \end{scope}
  \node[tag, anchor=west] at ([xshift=3mm,yshift=-1mm]trainingbox.north west)
    {Offline training};

  \node[input] (onlinecontext)
    at ($(context |- trainingbox.south)+(0,-14mm)$)
    {User/request features\\$\mathbf z_u,\mathbf z_{ur}$};
  \node[input, below=12mm of onlinecontext] (onlinepreds)
    {Selected upstream\\predicted scores $\mathbf v_{uri}$};
  \node[actor] (onlineactor) at (actor |- onlinecontext)
    {Actor\\$\mathbf w_{ur}=f_\theta(\mathbf z_u,\mathbf z_{ur})$};
  \node[proxy_score] (onlineproxy_score) at (proxy_score |- onlinecontext)
    {Proxy score\\$p_{uri}=\sum_{m=1}^M w_{ur}^m v_{uri}^m$};
  \node[rank_score, minimum width=37mm] (onlinerankscore)
    at (proxy_metric |- onlinecontext)
    {Multi-objective fusion formula\\$s_{uri}\leftarrow s_{uri}+\lambda\log(\max\{p_{uri},\epsilon\})$};
  \node[ranking_output] (ranking) at (critic |- onlinecontext)
    {Item ranking};

  \draw[flow] (onlinecontext) -- (onlineactor);
  \draw[flow] (onlineactor.east) -- (onlineproxy_score.west);
  \draw[flow] (onlinepreds.east) -| (onlineproxy_score.south);
  \draw[flow] (onlineproxy_score.east) -- (onlinerankscore.west);
  \draw[flow] (onlinerankscore.east) -- (ranking.west);

  \begin{scope}[on background layer]
    \node[group, fit=(onlinecontext)(onlinepreds)(onlineactor)
      (onlineproxy_score)(onlinerankscore)(ranking),
      inner sep=6mm] (servingbox) {};
  \end{scope}
  \node[tag, anchor=west] at ([xshift=3mm,yshift=-1mm]servingbox.north west)
    {Online serving};

  \draw[aux] (actor.south) -- (onlineactor.north);

  \end{tikzpicture}%
  }
  \caption{Overview of DCEO. During offline training, the actor $f_\theta$
  generates request-specific weights $\mathbf w_{ur}$ from the user and request
  features $\mathbf z_u$ and $\mathbf z_{ur}$. The weights combine the selected
  upstream predicted scores $\mathbf v_{uri}$ into the item-level proxy score
  $p_{uri}$, which is aggregated into the user-level proxy metric $P_u$. The
  critic $h_\phi$ models the ultimate objective $Y_u$ from $\mathbf z_u$ and
  $P_u$. The actor and critic are optimized using
  $\mathcal L_{\mathrm{actor}}$ and $\mathcal L_{\mathrm{critic}}$, respectively. During
  online serving, only $f_\theta$ is deployed, and $p_{uri}$ is added to the
  multi-objective fusion score $s_{uri}$ to improve the item ranking.}
  \Description{Overview of DCEO. During offline training, the actor $f_\theta$
  generates request-specific weights $\mathbf w_{ur}$ from the user and request
  features $\mathbf z_u$ and $\mathbf z_{ur}$. The weights combine the selected
  upstream predicted scores $\mathbf v_{uri}$ into the item-level proxy score
  $p_{uri}$, which is aggregated into the user-level proxy metric $P_u$. The
  critic $h_\phi$ models the ultimate objective $Y_u$ from $\mathbf z_u$ and
  $P_u$. The actor and critic are optimized using
  $\mathcal L_{\mathrm{actor}}$ and $\mathcal L_{\mathrm{critic}}$, respectively. During
  online serving, only $f_\theta$ is deployed, and $p_{uri}$ is added to the
  multi-objective fusion score $s_{uri}$ to improve the item ranking.}
  \label{fig:dceo_framework}
\end{figure*}
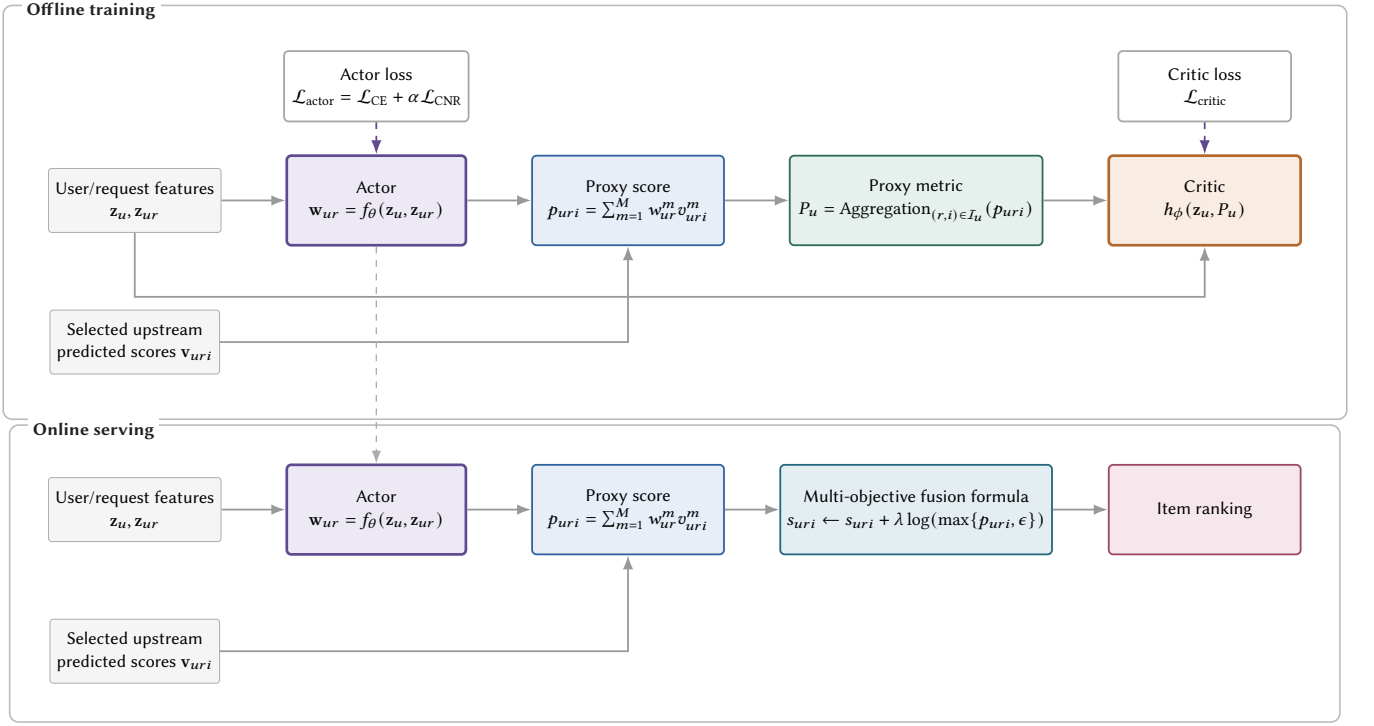

\section{Problem Formulation}
\label{sec:problem}

\subsection{User-Level Ultimate Objective and Granularity Gap}
\label{sec:granularity_gap}
Let $u$, $r$, and $i$ index a user, a search request from the user, and an item shown for that request, respectively. For each user $u$, we consider a reference day and define
\begin{equation}
    \mathcal I_u=\{(r,i): i \text{ is shown for request } r \text{ on the reference day}\}.
\end{equation}
Let $N_u=|\mathcal I_u|$ denote the corresponding impression count for user $u$.

Let $Y_u$ denote the user-level long-term objective accumulated over an $n$-day outcome window starting on the reference day. $Y_u$ may be defined as cumulative purchases, cumulative GMV, or another user-level long-term value metric.

Search ranking orders items within each request according to item-level scores, whereas $Y_u$ is a user-level quantity that summarizes the user's outcome over multiple requests. The granularity mismatch between item-level ranking scores and the user-level objective prevents $Y_u$ from directly serving as an item-level training label: all items exposed to the same user would otherwise receive the same label despite contributing differently to the outcome. Accordingly, we seek to learn item-level proxy scores that are better aligned with $Y_u$.

\subsection{Multi-Objective Fusion}
\label{sec:multi_objective_fusion}
For a request $r$ from user $u$, let $\mathcal C_{ur}$ denote the set of candidate items entering the ranking stage. For each item $i\in\mathcal C_{ur}$, the ranking stage provides $K$ predicted scores:
\begin{equation}
  \mathbf x_{uri}=[x_{uri}^{1},\ldots,x_{uri}^{K}]\in\mathbb R^K,
  \label{eq:fusion_predicted_scores}
\end{equation}
These predicted scores correspond to click, cart, purchase, transaction value, and other business objectives. The multi-objective fusion module combines them into a ranking score
\begin{equation}
  s_{uri}
  =\sum_{k=1}^{K}\lambda_k
  \log\!\left(\max\{x_{uri}^{k},\epsilon\}\right),
  \label{eq:production_fusion}
\end{equation}
where $\epsilon>0$ and each $\lambda_k$ is constant. Candidate items are ranked in descending order of $s_{uri}$.

\subsection{Item-Level Proxy Score and User-Level Proxy Metric}
\label{sec:contextual_proxy}
Consider an item-level proxy objective for the ultimate objective $Y_u$. Let $y_{uri}$ denote its label and $p_{uri}$ its predicted score. We assume that $p_{uri}$ is a calibrated prediction of $y_{uri}$, that is
\begin{equation}
  \mathbb E[p_{uri}]
  =\mathbb E[y_{uri}].
  \label{eq:proxy_calibration}
\end{equation}
A practical way to use this proxy objective to improve ranking is to add its predicted score to the multi-objective fusion formula:
\begin{equation}
  s_{uri}
  \leftarrow s_{uri}
  +\lambda\log\!\left(\max\{p_{uri},\epsilon\}\right),
  \label{eq:online_score}
\end{equation}
where $\lambda>0$ is a constant.

Empirically, adding a proxy score into the multi-objective fusion formula increases the action rate of the corresponding proxy objective. However, neither the daily action count nor the daily action rate is a suitable proxy metric because they both depend on the impression count, which is affected by ranking. For example, deboosting items with high predicted dislike scores may reduce the dislike rate while increasing user activity, thereby increasing the impression count. The daily dislike count may then increase despite the lower dislike rate. At a fixed impression count, however, the action count and action rate do not suffer from this problem. We therefore use the fixed-impression action rate as the user-level proxy metric. Under the calibration assumption, this rate can be expressed equivalently using the proxy score:

\begin{equation}
  \mathbb E_{(r,i)\in\mathcal{I}_u}[y_{uri}\mid N_u=C]
  =\mathbb E_{(r,i)\in\mathcal{I}_u}[p_{uri}\mid N_u=C].
  \label{eq:fixed_exposure_proxy_calibration}
\end{equation}

We thus define $P_u=\mathbb E_{(r,i)\in\mathcal{I}_u}[p_{uri}\mid N_u=C]$ as the user-level proxy metric. Accordingly, adding $p_{uri}$ to the fusion formula will increase $P_u$. The magnitude of this increase is controlled by $\lambda$, with a larger $\lambda$ producing a larger increase in $P_u$.

\subsection{Relative Causal Effect of the Proxy Metric on the Ultimate Objective}
\label{sec:problem_definition}
Suppose that adding the proxy score to the fusion formula increases $P_u$ by a relative amount $\delta>0$:
\begin{equation}
  P_u\longrightarrow(1+\delta)P_u.
  \label{eq:proxy_intervention}
\end{equation}
If the proxy metric is better aligned with the ultimate objective $Y_u$, the same relative increase in $P_u$ should produce a larger relative increase in $Y_u$. We use this relative increase in $P_u$ as the treatment and measure its effect on the ultimate objective. Let $Y_u(P)$ denote the ultimate objective when $P_u=P$. We define the relative causal effect as
\begin{equation}
  \mathrm{RCE} =
  \frac{\mathbb E_u[Y_u((1+\delta)P_u)-Y_u(P_u)]}
       {\mathbb E_u[Y_u(P_u)]}.
  \label{eq:rce}
\end{equation}
A larger positive $\mathrm{RCE}$ indicates better alignment of the proxy metric, and hence the underlying item-level proxy scores, with the ultimate objective. Accordingly, the learning objective is to find an item-level proxy score that maximizes $\mathrm{RCE}$.

\section{Method}
\label{sec:method}

\subsection{Overview}
\label{sec:framework_overview}
Figure~\ref{fig:dceo_framework} presents the training and serving framework of DCEO. Each training sample contains all impressions of a user on a reference day, together with the corresponding user-level features $\mathbf z_u$, request-level features $\mathbf z_{ur}$, selected upstream predicted scores $\mathbf v_{uri}$, and ultimate-objective label $Y_u$. The actor $f_\theta$ generates request-specific weights $\mathbf w_{ur}$ from $\mathbf z_u$ and $\mathbf z_{ur}$ and combines $\mathbf w_{ur}$ with $\mathbf v_{uri}$ to produce the item-level proxy score $p_{uri}(\theta)$. DCEO then aggregates the proxy scores over the user's impressions and uses the calibration model $g_\psi$ to calibrate the aggregate to a reference impression count, yielding the user-level proxy metric $P_u(\theta)$.

The critic $h_\phi$ estimates the ultimate objective $Y_u$ from $\mathbf z_u$ and $P_u(\theta)$. Its predictions before and after the relative intervention $P_u(\theta)\rightarrow(1+\delta)P_u(\theta)$ define the causal effect loss $\mathcal L_{\mathrm{CE}}$. The conditional normalized ranking loss $\mathcal L_{\mathrm{CNR}}$ further regularizes the actor by aligning the ordering of the proxy metric with that of the ultimate objective after conditional normalization. These two losses form the actor loss $\mathcal L_{\mathrm{actor}}$, while $g_\psi$, $h_\phi$, and the normalization model $e_\zeta$ are optimized with their corresponding losses. All four models are trained jointly.

During online serving, only the actor $f_\theta$ is deployed. It generates $\mathbf w_{ur}$ and computes $p_{uri}$ from the online user and request features and selected upstream predicted scores. The proxy score is added to the existing multi-objective fusion score $s_{uri}$ to improve the item ranking. We next describe each component in detail.

\subsection{User-Level Training Sample}
\label{sec:user_level_training_sample}
Each training sample contains all impressions of a user on a reference day. The inputs consist of user-level, request-level, and impression-level features. The user-level features $\mathbf z_u$ include the user's profile features and historical purchase sequence. They are stored only once in each sample because these features remain unchanged throughout the reference day. For each impression $(r,i)$, the request-level features $\mathbf z_{ur}$ include the query and contextual features, and the impression-level features include the selected upstream predicted scores $\mathbf v_{uri}\in\mathbb R^M$. The request-level and impression-level features vary across impressions, so each feature is stored as a sequence of length $N_u$.

Each sample is associated with multiple user-level outcome labels, including the cumulative click count, purchase count, and GMV over an $n$-day window starting on the reference day. One of these outcomes or a combination of them serves as the ultimate-objective label $Y_u$.

\subsection{Actor}
\label{sec:proxy_actor}
\subsubsection{Item-Level Proxy Score}
For each impression $(r,i)$, the actor model $f_\theta$ generates request-specific weights from the user features $\mathbf z_u$ and request features $\mathbf z_{ur}$
\begin{equation}
  \mathbf w_{ur}=f_\theta(\mathbf z_u,\mathbf z_{ur}), \qquad
  w_{ur}^m\geq 0, \qquad \sum_{m=1}^M w_{ur}^{m}=1.
  \label{eq:fusion_policy}
\end{equation}
The weights are then combined with the selected upstream predicted scores $\mathbf v_{uri}$ to produce the item-level proxy score
\begin{equation}
  p_{uri}(\theta)=\sum_{m=1}^{M}w_{ur}^{m}v_{uri}^{m}.
  \label{eq:problem_proxy_score}
\end{equation}

\subsubsection{User-Level Proxy Metric}
\label{sec:proxy_metric}
After calculating the item-level proxy scores $p_{uri}(\theta)$, we aggregate them into the user-level proxy metric $P_u(\theta)$. According to the definition in Section~\ref{sec:contextual_proxy}, we estimate $P_u(\theta)$ as
\begin{align}
  P_u^{\mathrm{raw}}(\theta)&=\frac{1}{N_u}\sum_{(r,i)\in\mathcal I_u}p_{uri}(\theta),
  \label{eq:raw_proxy_prediction}\\
  P_u(\theta)
  &=P_u^{\mathrm{raw}}(\theta) \notag\\
  &\quad {}\times\operatorname{sg}\!\left(
  \frac{g_\psi(\mathbf z_u,C)}
       {\max\{g_\psi(\mathbf z_u,N_u),\epsilon_g\}}
  \right),
  \label{eq:calibrated_sum}
\end{align}
where $C=100$ is the reference impression count and $g_\psi$ estimates the average of the item-level proxy scores $P_u^{\mathrm{raw}}(\theta)$ for user $u$ at a given impression count. The operator $\operatorname{sg}(\cdot)$ preserves its argument during the forward pass and blocks gradients through it during backpropagation. We train $g_\psi$ using a mean squared error loss
\begin{equation}
  \mathcal L_g=
  \mathbb E_u\!\left[
    \left(g_\psi(\mathbf z_u,N_u)
    -\operatorname{sg}\!\left(P_u^{\mathrm{raw}}(\theta)\right)\right)^2
  \right].
  \label{eq:calibration_loss}
\end{equation}
Equivalently, Eqs.~\eqref{eq:raw_proxy_prediction} and~\eqref{eq:calibrated_sum} first compute the average proxy score over the observed $N_u$ impressions and then multiply it by an impression-count calibration factor that maps the average at the observed count $N_u$ to its counterpart at the reference count $C$.

\subsection{Critic}
\label{sec:critic}
The critic model $h_\phi$ estimates the ultimate objective $Y_u$ for user $u$ given the user-level proxy metric $P_u(\theta)$. We train $h_\phi$ using a mean squared error loss
\begin{equation}
  \mathcal L_{\mathrm{critic}}=
  \mathbb E_u\left[(h_\phi(\mathbf z_u,\operatorname{sg}(P_u(\theta)))-Y_u)^2\right].
  \label{eq:critic_loss}
\end{equation}

\subsection{Optimization}
\label{sec:actor}
The actor is trained with two losses: the causal effect loss $\mathcal L_{\mathrm{CE}}$ and the conditional normalized ranking loss $\mathcal L_{\mathrm{CNR}}$.

\subsubsection{Causal Effect Loss}
We use the critic model to estimate the causal effect of increasing the user-level proxy metric from $P_u(\theta)$ to $(1+\delta)P_u(\theta)$ and use its negative as the causal effect loss
\begin{equation}
  \mathcal L_{\mathrm{CE}}=-\mathbb E_u[h_{\operatorname{sg}(\phi)}(\mathbf z_u,(1+\delta)P_u(\theta))-h_{\operatorname{sg}(\phi)}(\mathbf z_u,P_u(\theta))].
  \label{eq:ce_loss}
\end{equation}
When calculating $\mathcal L_{\mathrm{CE}}$, the parameters of the critic model $h_\phi$ are frozen so that $\mathcal L_{\mathrm{CE}}$ updates only the parameters of the actor model $f_\theta$.

The causal effect loss preserves the optimization direction of RCE while avoiding the instability of directly optimizing a ratio. The batch estimate of the baseline response $\mathbb E_u[h_\phi(\mathbf z_u,P_u(\theta))]$ varies little across training batches and can therefore be treated as approximately constant. Consequently, RCE is approximately proportional to $-\mathcal L_{\mathrm{CE}}$, so minimizing $\mathcal L_{\mathrm{CE}}$ approximately maximizes RCE. Directly using RCE as the actor loss would instead backpropagate through its denominator, making the gradients sensitive to fluctuations in the baseline response. We therefore optimize the unnormalized response difference in Eq.~\eqref{eq:ce_loss}.

\subsubsection{Conditional Normalized Ranking Loss}
The causal effect loss induces a highly non-convex optimization landscape for the actor, making it difficult to optimize effectively on its own. We therefore introduce a conditional normalized ranking loss as a regularizer. The goal of the conditional normalized ranking loss is to map all $(P_u(\theta),Y_u)$ pairs into the same distribution and then optimize their ordering. We first estimate the means $\mu_P$ and $\mu_Y$ and the log standard deviations $\log\sigma_P$ and $\log\sigma_Y$ of $P_u(\theta)$ and $Y_u$ using a model $e_\zeta$
\begin{equation}
  \left(\mu_P,\mu_Y,
  \log\sigma_P,\log\sigma_Y\right)
  =e_\zeta(\mathbf z_u).
  \label{eq:mean_logstd_prediction}
\end{equation}
$e_\zeta$ is trained by minimizing the Gaussian negative log-likelihood
\begin{equation}
  \begin{aligned}
  \mathcal L_e
  ={}&\mathbb E_u\!\Bigg[
    \log\sigma_P
    +\frac{\big(\operatorname{sg}(P_u(\theta))
      -\mu_P\big)^2}
           {2\sigma_P^2}
  \Bigg]\\
  &+\mathbb E_u\!\Bigg[
    \log\sigma_Y
    +\frac{\big(Y_u-\mu_Y\big)^2}
           {2\sigma_Y^2}
  \Bigg].
  \end{aligned}
  \label{eq:normalization_loss}
\end{equation}
We then define
\begin{equation}
  \begin{aligned}
  \widetilde P_u(\theta)
    &=\frac{P_u(\theta)-\operatorname{sg}(\mu_P)}
            {\max\{\operatorname{sg}(\sigma_P),\epsilon_e\}},\\
  \widetilde Y_u
    &=\frac{Y_u-\operatorname{sg}(\mu_Y)}
            {\max\{\operatorname{sg}(\sigma_Y),\epsilon_e\}}.
  \end{aligned}
\end{equation}
We compute the conditional normalized ranking loss from $\widetilde P_u(\theta)$ and $\widetilde Y_u$ using the Bradley--Terry loss:
\begin{equation}
  \mathcal L_{\mathrm{CNR}}=
  \frac{1}{|\mathcal B_{\mathrm{pair}}|}
  \sum_{(u,v)\in\mathcal B_{\mathrm{pair}}}
  \log\!\left(1+\exp\!\left[-(\widetilde P_u(\theta)-\widetilde P_v(\theta))\right]\right),
  \label{eq:cnr_loss}
\end{equation}
where $\mathcal B_{\mathrm{pair}}=\{(u,v):\widetilde Y_u>\widetilde Y_v\}$ denotes the set of ordered user pairs.

Intuitively, $\mathcal L_{\mathrm{CNR}}$ can be viewed as a coarse but stable approximation to $\mathcal L_{\mathrm{CE}}$. The causal effect loss implicitly groups users by their features $\mathbf z_u$ and, within each group, encourages $P_u$ to preserve the ordering of $Y_u$. Conditional normalization instead maps all $(P_u,Y_u)$ pairs to a common reference distribution, effectively placing them in a single group; the pairwise ranking loss then enforces the same ordering. Although this approximation discards some fine-grained conditioning on $\mathbf z_u$, the ranking signal is more stable than the difference between two critic predictions. Therefore, $\mathcal L_{\mathrm{CNR}}$ provides a stable complementary signal for optimizing the actor.

\subsubsection{Joint Training}
\label{sec:training}
The final loss for the actor model is
\begin{equation}
  \mathcal L_{\mathrm{actor}}=
  \mathcal L_{\mathrm{CE}}+\alpha\mathcal L_{\mathrm{CNR}},
  \label{eq:actor_loss}
\end{equation}
where $\alpha=0.3$ in the final configuration.

We jointly train the parameters of $f_\theta$, $g_\psi$, $h_\phi$, and $e_\zeta$ using the overall loss
\begin{equation}
  \mathcal{L}=\mathcal{L}_{\mathrm{actor}}+\mathcal{L}_g+\mathcal{L}_{\mathrm{critic}}+\mathcal{L}_e.
  \label{eq:dceo_loss}
\end{equation}

\subsection{Online Serving}
\label{sec:serving}
During online serving, only the actor model $f_\theta$ is deployed to compute request-specific weights
\begin{equation}
  \mathbf{w}_{ur}=f_\theta(\mathbf{z}_u,\mathbf{z}_{ur}).
  \label{eq:online_deployment_1}
\end{equation}
We then combine the selected upstream predicted scores $\mathbf v_{uri}$ with the weights $\mathbf w_{ur}$ to calculate the item-level proxy scores
\begin{equation}
  p_{uri}=\sum_{m=1}^M w_{ur}^m v_{uri}^m.
  \label{eq:online_deployment_2}
\end{equation}
We add $p_{uri}$ to the multi-objective fusion formula
\begin{equation}
  s_{uri}\leftarrow s_{uri}+\lambda\log\!\left(\max\{p_{uri},\epsilon\}\right).
\end{equation}

\section{Experiments}
\label{sec:experiments}

\subsection{Research Questions}
We evaluate DCEO by answering the following research questions:
\begin{itemize}
\item \textbf{RQ1:} How does DCEO perform under the final configuration?
\item \textbf{RQ2:} Is causal effect optimization better than predictive association optimization?
\item \textbf{RQ3:} How do the actor losses and upstream predicted-score set affect performance?
\item \textbf{RQ4:} How does the definition of the ultimate objective affect the learned weights?
\item \textbf{RQ5:} Does DCEO outperform a conventional GMV proxy in online search?
\end{itemize}

\subsection{Experimental Setup}
\label{sec:experimental_setup}

\subsubsection{Offline Data and Configuration}
We construct the user-level training samples described in Section~\ref{sec:user_level_training_sample} from search logs of a large-scale e-commerce search system. For each offline experiment, we train the model on the same 14 consecutive days of data and evaluate it on the following day. Unless otherwise specified, the ultimate objective is the cumulative GMV over four days starting on the reference day.

DCEO uses 17 selected upstream predicted scores listed in Table~\ref{tab:upstream_scores}. A simple transformation maps every predicted score into the interval $[0,1]$ while keeping its mean close to $0.1$, making the average actor weights comparable across scores. We set $C=100$, $\delta=0.05$, and $\alpha=0.3$ in the final configuration. Unless otherwise specified, we vary only the factor under study and keep all other settings fixed.

\begin{table}[t]
  \caption{Selected upstream predicted scores used by DCEO. These predicted scores are estimated by upstream models.}
  \label{tab:upstream_scores}
  \centering
  \scriptsize
  \begin{tabular}{@{}lp{0.62\columnwidth}@{}}
    \toprule
    Score name & Meaning \\
    \midrule
    \texttt{impr2click} & Probability of a click after an impression. \\
    \texttt{impr2cart} & Probability of a cart addition after an impression. \\
    \texttt{impr2pay} & Probability of a purchase after an impression. \\
    \texttt{impr2gmv} & GMV generated after an impression. \\
    \texttt{impr2pay10} & Probability of a purchase with transaction value above 10 after an impression. \\
    \texttt{impr2pay30} & Probability of a purchase with transaction value above 30 after an impression. \\
    \texttt{impr2pay100} & Probability of a purchase with transaction value above 100 after an impression. \\
    \texttt{impr2pay300} & Probability of a purchase with transaction value above 300 after an impression. \\
    \texttt{impr2pay1000} & Probability of a purchase with transaction value above 1,000 after an impression. \\
    \texttt{click2pay} & Probability of a purchase after a click. \\
    \texttt{click2gmv} & GMV generated after a click. \\
    \texttt{click2pay10} & Probability of a purchase with transaction value above 10 after a click. \\
    \texttt{click2pay30} & Probability of a purchase with transaction value above 30 after a click. \\
    \texttt{click2pay100} & Probability of a purchase with transaction value above 100 after a click. \\
    \texttt{click2pay300} & Probability of a purchase with transaction value above 300 after a click. \\
    \texttt{click2pay1000} & Probability of a purchase with transaction value above 1,000 after a click. \\
    \texttt{cart2gmv} & GMV generated after a cart addition. \\
    \bottomrule
  \end{tabular}
\end{table}

\subsubsection{Implementation Details}
DCEO does not rely on a specialized network architecture. All four models are multilayer perceptrons (MLPs) whose input features are represented as feature embeddings. The actor concatenates the user and request feature embeddings and uses a softmax output layer to produce the $M$ nonnegative weights in Eq.~\eqref{eq:fusion_policy}. The critic takes the user feature embeddings and the scalar proxy metric as input and produces a scalar prediction of the ultimate objective. The calibration model takes the user feature embeddings and impression count as input and outputs a scalar to estimate the average of the item-level proxy scores, while the normalization model takes the user feature embeddings as input and outputs the four conditional statistics in Eq.~\eqref{eq:mean_logstd_prediction}. All four models are trained jointly as described in Section~\ref{sec:training}.

\subsubsection{Offline Evaluation}
Our primary offline metric is the relative causal effect defined in Eq.~\eqref{eq:rce}, computed on the evaluation data as
\begin{equation}
  \mathrm{RCE}=
  \frac{\mathbb E_u[h_\phi(\mathbf z_u,1.05P_u)-h_\phi(\mathbf z_u,P_u)]}
       {\mathbb E_u[h_\phi(\mathbf z_u,P_u)]}.
  \label{eq:offline_rce}
\end{equation}
We report RCE as a ratio, so an RCE of $0.053$ corresponds to a $5.3\%$ relative increase in the ultimate objective under a $5\%$ increase in the proxy metric. To analyze the actor, we report the mean and standard deviation of each weight over all impressions. Because all predicted scores have the same range and a similar mean, their weight statistics are directly comparable. The weight mean reflects the importance of the corresponding predicted score in constructing the proxy score, while the weight standard deviation measures its personalization strength, with a larger value indicating stronger weight variation across user and request contexts.

\subsubsection{Online Evaluation}
We conduct a 41-day online A/B test in a large-scale e-commerce search system. The control group uses the conventional GMV proxy, defined as
\begin{equation}
  p_{\mathrm{GMV}}
  =\mathrm{pCTR}\times\mathrm{pCVR}
  \times\mathbb E[\text{transaction value}\mid\text{purchase}=1].
  \label{eq:conventional_gmv_proxy}
\end{equation}
The treatment group uses the DCEO proxy. Both proxies are incorporated into the existing multi-objective fusion formula using the same integration mechanism and boost strength. We report the relative changes of the treatment group over the control group in click count, purchase count, and GMV. GMV is the primary metric because it directly corresponds to the ultimate objective.

\subsection{Analysis of the Final Model (RQ1)}
\label{sec:final_model_analysis}
Under the final configuration, DCEO achieves an RCE of $0.053$. Table~\ref{tab:weight_distribution} summarizes the six predicted scores that receive most of the actor weight. \texttt{impr2click} receives the largest mean weight, while \texttt{click2pay10}, \texttt{click2pay30}, \texttt{click2pay100}, and \texttt{click2pay1000} collectively receive substantial weight.

All six predicted scores have nonzero weight standard deviations, showing that DCEO learns context-dependent weights rather than a single fixed combination.

\begin{table*}[t]
  \caption{Mean and standard deviation of representative actor weights in the final 4-day GMV model. The standard deviation measures personalization strength and is computed over impressions.}
  \label{tab:weight_distribution}
  \centering
  \small
  \begin{tabular}{lrrrrrr}
    \toprule
    Statistic & \texttt{impr2click} & \texttt{impr2pay} & \texttt{click2pay10} & \texttt{click2pay30} & \texttt{click2pay100} & \texttt{click2pay1000} \\
    \midrule
    Mean & 0.404 & 0.032 & 0.154 & 0.079 & 0.105 & 0.205 \\
    Standard deviation & 0.115 & 0.096 & 0.057 & 0.046 & 0.049 & 0.067 \\
    \bottomrule
  \end{tabular}
\end{table*}

\subsection{Causal Effect Optimization versus Predictive Association Optimization (RQ2)}
\label{sec:causal_vs_association}
Predictive association optimization trains the actor to maximize the ultimate objective predicted by the critic, $h_\phi(\mathbf z_u,P_u(\theta))$, and therefore favors proxy metrics associated with high ultimate-objective values. In contrast, causal effect optimization seeks a proxy metric whose increase produces a large estimated relative improvement in the ultimate objective. However, predictive association does not capture the effect of increasing the proxy metric on the ultimate objective. We therefore compare the two optimization paradigms.

We compare three actor losses: the predictive-association loss $-h_{\operatorname{sg}(\phi)}(\mathbf z_u,P_u(\theta))$, the causal effect loss $\mathcal L_{\mathrm{CE}}$, and the complete actor loss $\mathcal L_{\mathrm{actor}}$. All other configurations remain unchanged. We evaluate the three variants using RCE, which directly measures the estimated relative improvement in the ultimate objective under a $5\%$ increase in the proxy metric. As shown in Table~\ref{tab:causal_vs_association}, replacing predictive association optimization with $\mathcal L_{\mathrm{CE}}$ increases RCE from $0.022$ to $0.031$. Adding the conditional normalized ranking loss further increases RCE to $0.053$, a $2.41\times$ improvement over predictive association optimization. These results show that directly optimizing the causal effect better aligns changes in the proxy metric with changes in the ultimate objective.

\begin{table}[t]
  \caption{Comparison of predictive association and causal effect optimization. Higher RCE is better.}
  \label{tab:causal_vs_association}
  \centering
  \begin{tabular}{lc}
    \toprule
    Actor loss & RCE \\
    \midrule
    Predictive association optimization: $-h_{\operatorname{sg}(\phi)}(\mathbf z_u,P_u(\theta))$ & 0.022 \\
    Causal effect optimization: $\mathcal L_{\mathrm{CE}}$ & 0.031 \\
    Causal effect optimization: $\mathcal L_{\mathrm{actor}}$ & \textbf{0.053} \\
    \bottomrule
  \end{tabular}
\end{table}

\subsection{Component Ablation Study (RQ3)}
\label{sec:component_ablation}
\subsubsection{Effect of the Actor Losses}
We next study the contributions of $\mathcal L_{\mathrm{CE}}$ and $\mathcal L_{\mathrm{CNR}}$ and vary their trade-off coefficient $\alpha$. Table~\ref{tab:loss_ablation} shows that using $\mathcal L_{\mathrm{CE}}$ alone yields an RCE of $0.031$, while $\mathcal L_{\mathrm{CNR}}$ alone achieves an RCE of $0.048$. Combining the two losses further increases RCE, with $\alpha=0.3$ achieving the best result of $0.053$. A smaller coefficient of $0.1$ performs similarly, whereas increasing it to $1.0$ reduces RCE to $0.047$. These results indicate that the conditional normalized ranking loss provides useful regularization for actor learning, but overemphasizing it weakens RCE. We therefore use $\alpha=0.3$ in the final configuration.

\begin{table}[t]
  \caption{Ablation of the actor losses. Higher RCE is better.}
  \label{tab:loss_ablation}
  \centering
  \begin{tabular}{lc}
    \toprule
    Actor loss & RCE \\
    \midrule
    $\mathcal L_{\mathrm{CE}}$ & 0.031 \\
    $\mathcal L_{\mathrm{CNR}}$ & 0.048 \\
    $\mathcal L_{\mathrm{CE}}+0.1\mathcal L_{\mathrm{CNR}}$ & 0.052 \\
    $\mathcal L_{\mathrm{CE}}+0.3\mathcal L_{\mathrm{CNR}}$ & \textbf{0.053} \\
    $\mathcal L_{\mathrm{CE}}+1.0\mathcal L_{\mathrm{CNR}}$ & 0.047 \\
    \bottomrule
  \end{tabular}
\end{table}

\subsubsection{Effect of the Predicted-Score Set}
\label{sec:prediction_and_weight_analysis}
We compare five predicted-score sets that provide different types of information about the conversion process and transaction value. The GMV-only set uses \texttt{impr2gmv} as the item-level counterpart of the 4-day GMV objective. The basic set contains \texttt{impr2click}, \texttt{impr2cart}, \texttt{impr2pay}, and \texttt{impr2gmv}, representing the main outcomes after an impression. The value-aware set contains \texttt{impr2click}, \texttt{impr2cart}, \texttt{impr2pay}, \texttt{impr2pay10}, \texttt{click2pay30}, \texttt{click2pay100}, and \texttt{click2pay1000} to incorporate transaction-value information. The conversion-funnel set contains \texttt{impr2click}, \texttt{impr2cart}, \texttt{impr2pay}, \texttt{click2pay}, \texttt{click2gmv}, and \texttt{cart2gmv} to represent multiple stages of the conversion funnel. The full set combines all 17 predicted scores in Table~\ref{tab:upstream_scores}.

As shown in Table~\ref{tab:score_set}, the GMV-only set obtains an RCE of $0.027$, while the basic set improves it to $0.039$. The value-aware and conversion-funnel sets achieve RCEs of $0.041$ and $0.040$, respectively, and the full set achieves the highest RCE of $0.053$. These results show that \texttt{impr2gmv} alone is insufficient to construct a well-aligned proxy and suggest that predicted scores from different conversion stages and transaction-value ranges provide complementary information for learning a proxy that is better aligned with 4-day GMV.

\begin{table}[t]
  \caption{Effect of the upstream predicted-score set. Higher RCE is better.}
  \label{tab:score_set}
  \centering
  \begin{tabular}{lc}
    \toprule
    Predicted-score set & RCE \\
    \midrule
    \texttt{impr2gmv} only & 0.027 \\
    Basic impression-level set & 0.039 \\
    Value-aware set & 0.041 \\
    Conversion-funnel set & 0.040 \\
    Full 17-score set & \textbf{0.053} \\
    \bottomrule
  \end{tabular}
\end{table}

\begin{table*}[t]
  \caption{Mean actor weights under different ultimate objectives. The table reports representative predicted scores defined in Table~\ref{tab:upstream_scores}; the remaining predicted scores account for the unreported weight mass.}
  \label{tab:objective_weights}
  \centering
  \small
  \begin{tabular}{lrrrrrr}
    \toprule
    Ultimate objective & \texttt{impr2click} & \texttt{click2pay} & \texttt{click2pay10} & \texttt{click2pay30} & \texttt{click2pay100} & \texttt{click2pay1000} \\
    \midrule
    4-day click count & 0.999 & 0.000 & 0.000 & 0.000 & 0.000 & 0.000 \\
    4-day purchase count & 0.531 & 0.101 & 0.256 & 0.000 & 0.000 & 0.000 \\
    4-day GMV & 0.404 & 0.032 & 0.154 & 0.079 & 0.105 & 0.205 \\
    4-day purchase count $+\,0.1\times$ GMV & 0.407 & 0.000 & 0.056 & 0.098 & 0.103 & 0.187 \\
    \bottomrule
  \end{tabular}
\end{table*}

\begin{table*}[t]
  \caption{Mean actor weights for GMV objectives with different accumulation windows. The table reports representative predicted scores defined in Table~\ref{tab:upstream_scores}; the remaining predicted scores account for the unreported weight mass.}
  \label{tab:horizon_weights}
  \centering
  \small
  \begin{tabular}{lrrrrrr}
    \toprule
    Ultimate objective & \texttt{impr2click} & \texttt{click2pay} & \texttt{click2pay10} & \texttt{click2pay30} & \texttt{click2pay100} & \texttt{click2pay1000} \\
    \midrule
    1-day GMV & 0.331 & 0.034 & 0.108 & 0.144 & 0.127 & 0.158 \\
    2-day GMV & 0.379 & 0.000 & 0.000 & 0.140 & 0.129 & 0.189 \\
    3-day GMV & 0.417 & 0.000 & 0.155 & 0.081 & 0.122 & 0.202 \\
    4-day GMV & 0.404 & 0.032 & 0.154 & 0.079 & 0.105 & 0.205 \\
    \bottomrule
  \end{tabular}
\end{table*}

\subsection{Effect of the Ultimate Objective (RQ4)}
\label{sec:ultimate_objective_analysis}
\subsubsection{Effect of the Objective Type}
We train DCEO with four user-level ultimate objectives while keeping the other settings fixed. Table~\ref{tab:objective_weights} reports the mean weights of six representative predicted scores. When the ultimate objective is 4-day click count, the actor assigns nearly all weight to \texttt{impr2click}. For 4-day purchase count, it retains a large \texttt{impr2click} weight but shifts substantial mass to \texttt{click2pay} and \texttt{click2pay10}. When optimizing 4-day GMV, the actor distributes more weight across transaction-value thresholds, including a weight of $0.205$ on \texttt{click2pay1000}. The composite purchase-and-GMV objective produces a related but distinct allocation. These changes show that DCEO does not learn a fixed fusion rule: its proxy composition responds to the specified user-level objective.

\subsubsection{Effect of the Objective Horizon}
We further fix the ultimate objective to GMV and vary its accumulation window from 1 to 4 days. Table~\ref{tab:horizon_weights} reports the resulting mean actor weights. The \texttt{impr2click} weight increases from $0.331$ for 1-day GMV to $0.404$ for 4-day GMV. A larger \texttt{impr2click} weight gives greater preference to items with high \texttt{impr2click} scores and therefore tends to increase click count. A higher click count indicates that users explore more items. This pattern suggests that DCEO encourages more exploration when optimizing GMV over a longer horizon and provides further evidence that DCEO adapts the composition of its proxy score to the definition of the ultimate objective.

\subsection{Online A/B Test (RQ5)}
\label{sec:online_ab}
Table~\ref{tab:online_results} presents the online A/B test results. Compared with the conventional GMV proxy, DCEO increases GMV by $0.36\%$. It also increases click count and purchase count by $0.36\%$ and $0.12\%$, respectively. The improvement in GMV demonstrates that DCEO is more effective than the conventional GMV proxy in optimizing the ultimate objective. The increases in click count and purchase count show that the GMV gain does not come at the cost of fewer clicks or purchases.

\begin{table}[t]
  \caption{Online relative changes of DCEO over the conventional GMV proxy in the 41-day A/B test.}
  \label{tab:online_results}
  \centering
  \begin{tabular}{lr}
    \toprule
    Metric & Relative change \\
    \midrule
    Click count & $+0.36\%$ \\
    Purchase count & $+0.12\%$ \\
    GMV & $\mathbf{+0.36\%}$ \\
    \bottomrule
  \end{tabular}
\end{table}

\section{Limitations}
DCEO has several limitations. First, RCE is a model-based local effect estimate obtained from a critic trained on observational logs. Its causal interpretation requires the user features to capture the major confounders between the proxy metric and the ultimate objective, sufficient data support for both $P_u$ and $(1+\delta)P_u$, and accurate critic predictions within this local region. Unobserved confounding may bias causal identification, while insufficient support or critic misspecification may introduce estimation error. We use a small intervention magnitude of $\delta=0.05$ to reduce local extrapolation, although this does not eliminate bias from unobserved confounding. The online A/B test validates the end-to-end effectiveness of the learned proxy score but does not directly validate the numerical RCE estimate. Future work could randomly vary the proxy-score boost strength $\lambda$ to collect interventional data and more directly estimate the response of the ultimate objective to changes in the proxy metric.

Second, calibrating the proxy metric to a fixed impression count makes action counts and rates comparable across users. This design, however, deliberately excludes changes in the ultimate objective that arise through ranking-induced changes in user activity and impression count. The offline RCE should therefore be interpreted as an alignment metric at the reference impression count rather than an estimate of the total deployment effect. The online A/B test complements this metric by measuring the end-to-end impact of DCEO, including changes in user activity and impression count. Future work could jointly model the per-impression response and the impression count to capture both pathways offline.

Third, a more general actor could directly map impression-level features to an item-level proxy score. However, the resulting function space was difficult to optimize stably under our current critic-based objective. We therefore restrict the actor to predicting context-dependent weights that form a convex combination of selected upstream predicted scores. Compared with direct proxy-score prediction, this parameterization improves optimization stability and enables interpretable, lightweight online serving. The trade-off is reduced expressiveness: the actor cannot recover information absent from the upstream scores and inherits limitations in their coverage and quality. Future work will improve optimization stability to support more expressive actors that directly use impression-level features.

Finally, our evaluation is limited to one e-commerce search system and objective horizons of up to four days. Validation on additional platforms, objectives, and longer horizons is needed to establish broader generalizability.

\section{Conclusion}
\label{sec:conclusion}
In this paper, we presented DCEO, a data-driven framework for addressing the granularity gap between item-level ranking and user-level long-term objectives in industrial e-commerce search. DCEO learns context-dependent item-level proxy scores, aggregates them into a user-level proxy metric, and uses an actor-critic framework to directly optimize the relative causal effect of the proxy metric on the ultimate objective. During online serving, only the actor is deployed, and the learned proxy score is added to the existing multi-objective fusion formula. Extensive offline experiments and analyses demonstrate the effectiveness and interpretability of DCEO. In a 41-day online A/B test, DCEO outperforms the conventional GMV proxy by $0.36\%$ in GMV. These results demonstrate the effectiveness of learning item-level ranking signals directly from user-level long-term supervision.

\section{Ethical Considerations}
This study uses behavioral logs and user-side features from an industrial e-commerce search system, which may contain sensitive information about users' interests and purchasing behavior. Responsible use requires appropriate data access controls, minimization, de-identification, and aggregation. We report only aggregate experimental results. DCEO may inherit biases from historical interactions, upstream models, and the existing ranking system, and may change the exposure of different items and merchants. Deployments should therefore evaluate relevant user, item, and merchant groups and monitor fairness, diversity, and user experience. Online A/B tests should use staged rollout, continuous monitoring, and rollback mechanisms. Since the impact of DCEO depends on the chosen objective, objective selection requires human oversight and consideration of affected stakeholders.

\bibliographystyle{ACM-Reference-Format}
\bibliography{sample-base}



\end{document}